# Quantitative Technology Forecasting: a Review of Trend Extrapolation Methods

Peng-Hung Tsai*[1], Daniel Berleant*[1], Richard S. Segall[2], Hyacinthe Aboudja[3], Venkata Jaipal R. Batthula[1], Sheela Duggirala[1] and Michael Howell[1]

*ptsai@ualr.edu, jdberleant@ualr.edu; [1]Dept. of Information Science, University of Arkansas at Little Rock; [2]Dept. of Information Systems & Business Analytics, Arkansas State University, Jonesboro; [3]Dept. of Computer Science, Oklahoma City University

## Abstract

Quantitative technology forecasting uses quantitative methods to understand and project technological changes. It is a broad field encompassing many different techniques and has been applied to a vast range of technologies. A widely used approach in this field is trend extrapolation. Based on the publications available to us, there has been little or no attempt made to systematically review the empirical evidence on quantitative trend extrapolation techniques. This study attempts to close this gap by conducting a systematic review of technology forecasting literature addressing the application of quantitative trend extrapolation techniques. We identified 25 studies relevant to the objective of this research and classified the techniques used in the studies into different categories, among which growth curves and time series methods were shown to remain popular over the past decade, while newer methods, such as machine learning-based hybrid models, have emerged in recent years. As more effort and evidence are needed to determine if hybrid models are superior to traditional methods, we expect to see a growing trend in the development and application of hybrid models to technology forecasting.

*Keywords*: technological forecasting; trend extrapolation; technology evolution; systematic literature review

## 1. Introduction

For many years, researchers have investigated how to measure technology change and forecast its development. As a result, numerous technology forecasting techniques have been studied and applied to a variety of technologies. Nevertheless, there is not yet a generally accepted method superior to any alternatives, as the choice of which method to use is affected by the availability of data and experts, the field in which the predictions are applied, and the needs of the intended users (National Research Council, 2010). In general, the developed methods can be divided into qualitative and quantitative ones (Chou, 2011; Coates et al., 2001; Lee, 2021; Batthula et al., 2021). Qualitative methods use the intuition, expertise, and insight of experts to understand and forecast technological development. A popular example in this class is the Delphi method (Chou, 2011; Coates et al., 2001), which was proposed in the 1950s by RAND Corporation to forecast the impact of technology on warfare (Cho and Daim, 2013). It uses a questionnaire to obtain opinions about a topic from a panel of experts with the aim of achieving a well-informed consensus on future prospects of a technology. Quantitative methods, on the other hand, use historical quantitative data to identify technology change patterns in the data that can be extended into the future to obtain a forecast. In this category, a widely used approach is quantitative trend extrapolation. That approach first employs curve fitting to derive a trend line or curve that best fits a given set of historical data points and then extrapolates the resulting line or curve into the future to generate forecasts, yielding a trend extrapolation. Forecasting success in this approach is dependent on selecting a suitable fitting curve (Calleja-Sanz et al., 2020).

In this study, we aim to explore the technology forecasting literature on quantitative trend extrapolation techniques and provide an overview of the present state of research regarding the methods and their applications in different technological domains. To achieve this aim, we employ the systematic literature review (SLR) method. Given the huge and ever-growing body of literature available, a literature review that includes only a few studies, which may or may not be representative of all relevant studies, is subject to significant biases. On the other hand, by adhering to certain guidelines, SLRs provide an effective method for minimizing bias in the selection of studies as well as ensuring an objective review of the literature. Furthermore, compared to traditional (non-systematic) literature reviews, SLRs allow researchers to research the literature of a topic area in a transparent and rigorous way (Kitchenham and Charters, 2007). As SLRs aim to comprehensively collect related research on a topic in an unbiased and reliable manner (Shaffril et al., 2021), it is required to create a review protocol at the very beginning that clearly describes the research question to be addressed and details of the methods to be used to conduct the review. Additionally, selection criteria and search strategies must be defined for screening the studies and gathering as much of the relevant literature as possible. It is also required to document all pre-defined guidelines (such as search strategies, selection criteria, etc.), list the databases searched, and give the methods used, thereby ensuring transparency, validity, and replicability. It is the rigor involved in the search process that distinguishes SLRs from traditional reviews.

A number of studies have surveyed the literature on technology forecasting. Petrina (1990) reviewed the historical backgrounds and methods of technological forecasting and technology assessment and compared their similarities and differences. Coates et al. (2001) looked at the evolution of technological forecasting and discussed its social, political, business, and institutional drivers. Oliver et al. (2002) undertook a preliminary survey of long-term technological forecasting techniques directed to military needs. In addition, Martino (2003) described developments in technological forecasting methodology including environmental scanning, models, scenarios, Delphi, extrapolation, probabilistic forecasts, technology measurement, and chaos theory. Yu (2007) reviewed and compared major technology forecasting methods classified by the specific underlying assumptions about how the technology innovation process works to determine if the decision-focused scenario approach is useful for technology forecasting in a volatile business environment. Firat et al. (2008) summarized the field of technological forecasting, covering its evolution, techniques, and applications, and analyzed how it is used in practice. In the past decade other work has also appeared. A study conducted by Cho and Daim (2013) provided a historical overview of technology forecasting as well as an examination of the characteristics and chronological evolution of various technology forecasting methods. Kang et al. (2013) tracked the evolution of technology forecasting research activities and looked at the relationship between the forecasting approaches used and the industries using the approaches. Meade and Islam (2015) reviewed the approaches and applications of diffusion modeling and forecasting, time series forecasting, and technological forecasting used in the information and communications technology area. Moreover, Doos et al. (2016) carried out a systematic review of the methods used to predict future health technologies within a span of 3–20 years. Haleem et al. (2018) examined the historical evolution of technology forecasting and assessment and identified its methodologies along with the analysis of the relationship among the identified methods. Similarly, Calleja-Sanz et al. (2020) explored the evolution of technology forecasting methods and discussed their concepts, developments, and recent trends. More recently, Lee (2021) presented an overview of journal articles on data analytics in technology forecasting published in the field of technology and innovation management, and Viet et al. (2021) systematically reviewed data mining methods for emerging technology trend analysis and forecasting. Table 1 presents a comparison of existing surveys of the literature on technology forecasting (TF). In summary, this body of work has the following limitations.

- Most surveys did not use the SLR approach to conduct the review.

- Most surveys are not up to date, and thus many relevant studies published in recent years have been missed.

- Several studies concentrated on a specific technological domain where the TF methods were applied, thus excluding consideration of the field overall.

- None of the studies comprehensively and thoroughly addressed the quantitative techniques for trend extrapolation applied to diverse contexts.

By addressing these weaknesses, this study offers three major contributions to the scientific literature.

- It is the first attempt, to the best of the authors' knowledge, to conduct an SLR on technology forecasting addressing the application of quantitative trend extrapolation techniques to various technologies.

- It gives an updated picture of the published studies on applying quantitative trend extrapolation to technology forecasting and serves as a reference for technology forecasting researchers interested in employing the SLR approach.

- It helps familiarize scholars and practitioners with the current state of quantitative trend extrapolation methods for technology forecasting, better enabling them to identify directions for developing and improving the methods.

**Table 1:** Comparison of existing surveys of the technology forecasting (TF) literature

| Authors | Strengths | Limitations |
|---|---|---|
| Petrina (1990) | • Discussed the history and evolution of TF.<br>• Classified TF methods and divided them into component methods. | • Provided an overview of TF methods without delving into each one in depth.<br>• Lacked a picture of the current state of TF research related to quantitative trend extrapolation.<br>• Not an SLR study. |
| Coates et al. (2001) | • Examined TF's past and provided detailed insights into its future.<br>• Classified TF methods into different groups. | • Trend extrapolation techniques received little discussion.<br>• Lacked a picture of the current state of TF research related to quantitative trend extrapolation.<br>• Not an SLR study. |
| Oliver et al. (2002) | • Divided TF methods into formal and informal approaches for predicting technologies applicable to the military.<br>• Provided comparisons among the methods for best applications and suitable time frames. | • Trend extrapolation techniques received little discussion.<br>• Lacked a picture of the current state of TF research related to quantitative trend extrapolation.<br>• Not an SLR study. |
| Martino (2003) | • Discussed circa 2003 recent developments and new approaches to technological forecasting. | • The discussion of trend extrapolation techniques was centered on growth curves.<br>• Lacked a picture of the current state of TF research related to quantitative trend extrapolation.<br>• Not an SLR study. |
| Yu (2007) | • Compared major TF methods and described their advantages, pitfalls, and applicabilities. | • Trend extrapolation techniques received little discussion.<br>• Lacked a picture of the current state of TF research related to quantitative trend extrapolation.<br>• Not an SLR study. |

| Authors | Strengths | Limitations |
|---------|-----------|-------------|
| Firat et al. (2008) | • Presented a summary of some of the most popular TF methods and discussed their use in business | • There was not much discussion of trend extrapolation methods<br>• Lacked a picture of the current state of TF research related to quantitative trend extrapolation<br>• Not an SLR study |
| Cho and Daim (2013) | • Introduced TF from a historical perspective<br>• Reviewed techniques that combine different TF methods to improve forecasting accuracy | • There was not much discussion of trend extrapolation methods<br>• Lacked a picture of the current state of TF research related to quantitative trend extrapolation<br>• Not an SLR study |
| Kang et al. (2013) | • Extensively collected the TF studies for the time period beginning with the first published journal article<br>• Identified development trends and patterns of TF research and application areas of TF methods | • Trend extrapolation techniques received little discussion<br>• Lacked a picture of the current state of TF research related to quantitative trend extrapolation<br>• Not an SLR study |
| Meade and Islam (2015) | • Reviewed the literature according to applications and forecasting approaches in the area of information and communications technology (ICT) | • The study context where TF methods were applied was limited to the domain of ICT<br>• The TF research activities in the context of ICT were limited and mostly expert opinion based<br>• Lacked a picture of the current state of TF research related to quantitative trend extrapolation<br>• Not an SLR study |

| Authors | Strengths | Limitations |
| --- | --- | --- |
| Doos et al. (2016) | • Conducted a systematic search of the literature to collect studies<br>• Provided a comprehensive review of the approaches used to forecast health technologies and the types of technologies targeted | • The study context where TF methods were applied was limited to the domain of health technologies<br>• The methods used by the selected studies were mainly the approaches involving surveys of experts, with the absence of techniques for trend extrapolation<br>• Lacked a picture of the current state of TF research related to quantitative trend extrapolation |
| Haleem et al. (2018) | • Conducted an extensive search of the literature spanning 1960-2015<br>• Summarized the TF methods, the contexts in which the methods are applied, and the limitations of the methods | • Fitted the identified TF methods into different families without giving many details about each method<br>• Trend extrapolation techniques received little discussion<br>• Lacked a picture of the current state of TF research related to quantitative trend extrapolation<br>• Not an SLR study |
| Calleja-Sanz et al. (2020) | • Divided TF methods into different groups and discussed in detail the concepts, recent advances, and new methods for each group | • The discussion of trend extrapolation techniques was limited<br>• Lacked a picture of the current state of TF research related to quantitative trend extrapolation<br>• Not an SLR study |
| Lee (2021) | • Focused on reviewing the literature applying quantitative methods and models to TF<br>• Identified relevant TF studies with an established framework | • Lacked specific details involved in the methods surveyed<br>• Trend extrapolation techniques was little discussed<br>• Not an SLR study |
| Viet et al. (2021) | • Investigated the literature employing data mining methods for forecasting emerging technologies based on the systematic mapping methodology | • Lacked specific details involved in the methods surveyed<br>• The discussion of trend extrapolation techniques was limited<br>• Not an SLR study |

## 2. Methodology

We followed a five-step process for SLRs as proposed by Khan et al. (2003). We detail these steps in the following subsections.

### 2.1 Framing questions for a review

To present an up-to-date overview of research on applying quantitative trend extrapolation to technology forecasting, we seek to address the following key questions.

Q1: What does existing research provide about the implementation of quantitative trend extrapolation methods for technology forecasting?

Q2: What are the quantitative trend extrapolation methods that have been adopted and what are the technologies to which they have been applied?

### 2.2 Identifying relevant work

A review protocol was developed to identify and choose relevant articles using the PRISMA (Preferred Reporting Items for Systematic Reviews and Meta-Analyses) statement by Moher et al. (2009) as a guide. The PRISMA methodology is the acknowledged standard for performing an SLR. Figure 1 shows a flow diagram of the search strategy followed by details of the search and selection process.

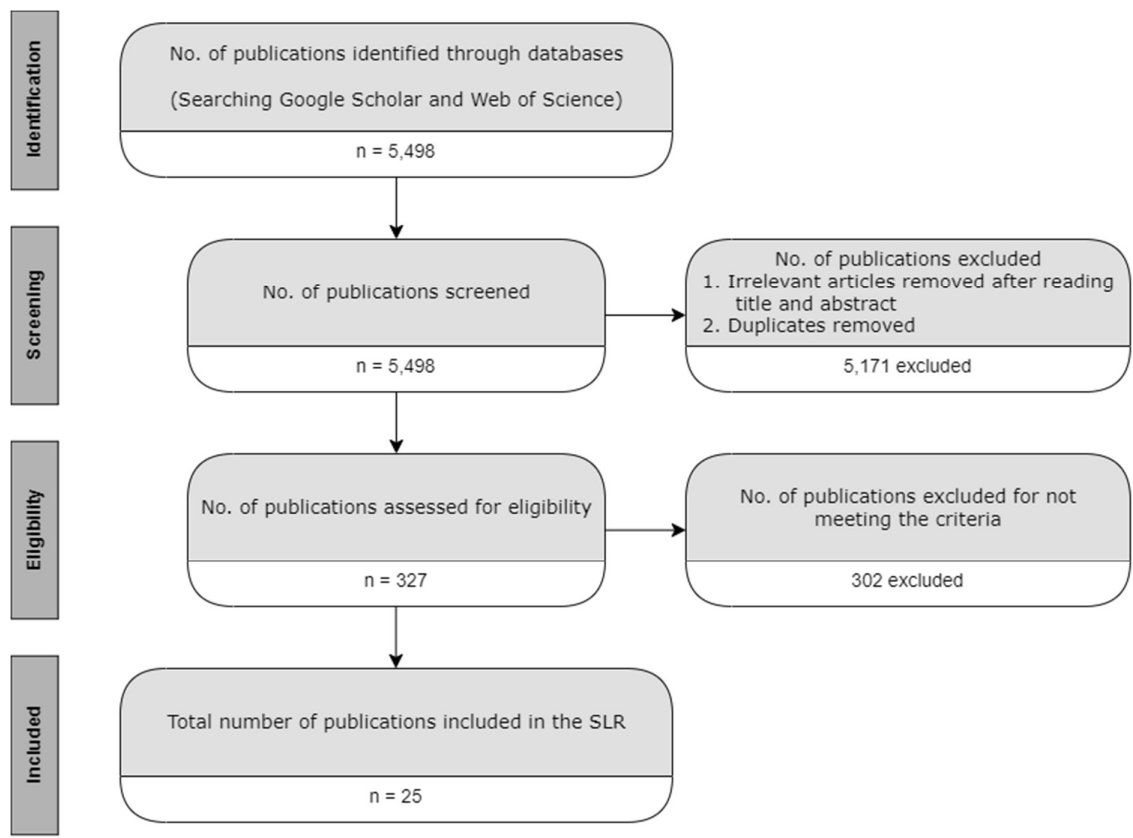

**Figure 1:** Flow diagram for the search and selection process

The steps in the process are as follows:

- **Identification**

We chose Google Scholar and Web of Science to search for the relevant publications. Google Scholar was used because it offers relatively extensive coverage of academic publications compared to other online literature databases (Martín-Martín, 2018) and is freely accessible. Web of Science, which is one of the most recognized databases in many scientific fields (Chadegani et al., 2013), was used in order to broaden the list of collections and provide a crosscheck.

The search consisted of search queries for each database suited to the technical and terminological similarities and differences between these two databases. We performed the Google Scholar search on October 12, 2022, with two different queries. We first broadly searched full text articles with specified keywords such as technology, forecasting, progress, quantitative, machine learning, extrapolate, error, and their synonyms and related words, occurring anywhere in the article (see Table 2). Then we limited the search to titles. We left the "include patents" and "include citations" boxes unchecked to exclude patents and inaccessible publications. We searched the Web of Science platform on the same date, with three different queries. First was a search of All Databases, which provides cross-search of multiple databases, with the first query executed on title. Second was a search of All Databases again with the second query executed on topic. Third was a search of the Web of Science Core Collection with the third query executed on all fields. Each search query consisted of different keywords combined with the appropriate Boolean operators, quotation marks, and/or the asterisk truncation symbol *. While searching both Google Scholar and Web of Science, we set the range of publication dates from 1990 to 2022 to keep our review to manageable levels. The beginning year 1990 was chosen as we considered that Machine Learning, an increasingly popular approach for curve fitting and trend extrapolation, began to take off in the 1990s. Also, we limited the search results to English language articles.

After our database search, an initial 5,498 articles were identified, of which 4,536 were found first in Google Scholar and an additional 962 in Web of Science. Table 2 shows, after a series of keyword screenings, the ultimately used query strings and the corresponding search.

**Table 2:** Search strings and results

| Database | Query | Search string | Results |
|---|---|---|---|
| Google Scholar | 1 | ("technology\|technological forecasting\|foresight\|prediction") ("technology\|technological progress\|progression\|advancement\|development\|evolution\|change\|growth") (quantitative\|mathematical\|statistical\|"machine learning") extrapolate error\|residual | 2,097 |
| | 2 | allintitle: technology\|technological\|technologies forecasting\|forecast\|forecasts | 2,439 |
| Web of Science | 1 | (TI=(technolog* forecast*)) AND (LA==("ENGLISH")) | 600 |
| | 2 | (TS=("technology forecasting" OR "technological forecasting" OR "technology foresight" OR "technological foresight" OR "technology prediction" OR "technological prediction")) AND TS=(quantitative OR mathematical OR statistical OR "machine learning") and English (Languages) | 268 |
| | 3 | ((ALL=("technology forecasting" OR "technological forecasting" OR "technology foresight" OR "technological foresight" OR "technology prediction" OR "technological prediction")) AND ALL=("technology progress" OR "technological progress" OR "technology progression" OR "technological progression" OR "technology advancement" OR "technological advancement" OR "technology development" OR "technological development" OR "technology evolution" OR "technological evolution" OR "technology change" OR "technological change" OR "technology growth" OR "technological growth")) AND ALL=(quantitative OR mathematical OR statistical OR "machine learning") and English (Languages) | 94 |
| Total | | | 5,498 |

- **Screening**

Next, we manually screened the identified literature at the level of titles and abstracts. The purpose was to narrow down the number of articles for in-depth review of their full texts. To achieve this end, we applied the inclusion and exclusion criteria as stated to select the articles.

- **Inclusion Criteria:**

1. Articles involving using quantitative trend extrapolation techniques for technology forecasting.

2. Articles such as journal articles, conference papers, and other forms of articles exclusive of theses and book chapters.

3. Articles published during the period of 1990-2022.

- **Exclusion Criteria:**

1. Articles dealing with technology diffusion or adoption, which is the process by which a technology is adopted by a population (Meade and Islam, 2015) and thus does not suit our purpose of analyzing trends in technological advancement and projecting the future technology performance.

2. Theoretical or concept articles not using a curve fitting method to fit a particular dataset.

3. Articles not written in English.

4. Duplicate articles.

5,171 articles were filtered out by the inclusion and exclusion criteria of the screening step. After excluding them, we had 327 articles ready for further examination.

- **Eligibility:**

Subsequently, we surveyed the full texts of the remaining 327 articles and used the following criteria to determine their eligibility.

1. Articles must clearly describe the forecasting techniques utilized and the technologies where the techniques were applied.

2. Articles must explicitly present the data sources and measures/indicators used for measuring the technological performance.

3. Articles must assess the forecasting performance of the utilized techniques using a hold-out sample and an error measure so that technique credibility and adequacy can be demonstrated.

The full text review filtered out 302 of the 327 remaining articles, giving a final collection of 25 articles eligible to be considered for detailed analysis.

## 2.3 Assessing the quality of studies

The eligible articles were validated to finalize the selection of articles to be included in the analysis. To do this, the primary author verified the selected articles for quality and relevance by re-examining the entire content of each article according to the inclusion, exclusion, and eligibility criteria and taking into account the relevance of each article to each pre-defined research question. Then, the entire set of selected articles was reviewed again by the second author independently against the same criteria to avoid bias in selection. After the validation process, it was confirmed that all 25 articles met the criteria and were qualified to be included in the detailed analysis.

## 2.4 Summarizing the evidence

We aggregated and summarized the data extracted from the selected articles to facilitate interpreting our findings, including details on each method used, in the next section. Table 3 surveys the contexts of the articles included in the SLR, describing the authors, country of the study, types of technologies targeted, methods used, and sources of data. It highlights that most of the studies have more than one author and among them, a few studies were conducted by the researchers from multiple countries, signaling ongoing international collaboration efforts in this field. The numbers of papers in the included studies from different countries is presented in Figure 2. As it shows, the United States stands out among the total of 12 countries with 14 papers. The next major source of papers was the United Kingdom with 4 papers, followed by mainland China with 3 papers.

**Table 3:** Some key properties of the articles included in this review

| Authors | Country | Technology | Trend extrapolation methods | Data source |
|---|---|---|---|---|
| Inman et al. (2005) | US | U.S. fighter jets | Linear regression | Dataset from a previous paper |
| Inman et al. (2006) | US | U.S. fighter jets | Linear regression | Dataset from a previous paper |
| Yoon and Lee (2008) | South Korea, UK | Various technologies by industry ( i.e., bio-technology, information technology, etc.) | Bass model (S-shaped curve) | The United States Patent and Trademark Office (USPTO) website database |
| Lamb et al. (2010) | US | Commercial airplane | Linear regression, logistic (Pearl) curve (S-shaped curve) | Datasets from company websites, reports, and aircraft interest websites |
| Widodo et al. (2011) | Indonesia | Biomedical technology | Linear regression, polynomial regression, Holt-Winters exponential smoothing (Winter's exponential smoothing), Holt's exponential smoothing, support vector regression (SVR) | PubMed website database |
| Yuan et al. (2011) | Taiwan | Organic light emitting diodes (OLEDs) | Bass model (S-shaped curve), Gompertz curve (S-shaped curve), GM(1,1) grey model | Historical annual power efficiency of OLED data |
| Bailey et al. (2012) | US | Sixty-two different technologies | Moore's law (exponential growth law), Wright's law (experience curve), Goddard's law (experience curve), Moore's law random walk, Goddard's law random walk, Moore-Goddard's law random walk | Performance Curve Database at http://pcdb.santafe.edu |

| Authors | Country | Technology | Trend extrapolation methods | Data source |
|---------|---------|------------|----------------------------|-------------|
| Lim et al. (2012) | US | Wireless communication | Linear regression | Datasets from previous studies |
| Sood et al. (2012) | US | 25 different technologies selected from six distinct markets: automotive motive battery technologies, data transfer, desktop printers, display monitors, desktop memory, and external lighting technologies | Bass model (S-shaped curve), Gompertz curve (S-shaped curve), Kryder's law (exponential growth law), logistic (Pearl) curve (S-shaped curve), Moore's law (exponential growth law), diff reg method (linear regression), Gupta model, step and wait (SAW) model, tobit II model | Primarily from white papers, technical journals, industry reports, museum records, press announcements, and main companies' timelines. |
| Smith and Agrawal (2015) | US | Data processing technology, including three classes, which are artificial intelligence (AI), database and file management or data structures (DFMDS), and generic control systems or specific applications (GCSSA). | Autoregressive integrated moving averages (ARIMA), Holt-Winters exponential smoothing | UC Berkley Fung Institute patent database |
| Farmer and Lafond (2016) | UK | 53 different technologies | Geometric random walk with drift | The Santa-Fe Performance Curve DataBase |
| Nagula (2016) | India | Fuel cell technology in hybrid and electric vehicles | Gompertz curve (S-shaped curve), logistic (Pearl) curve (S-shaped curve), Weibull model (S-shaped curve) | The United States Patent and Trademark Office (USPTO) website database |
| Cho and Anderson (2017) | US | Main battle tank (MBT) | Linear regression | Dataset from a previous paper |

| Authors | Country | Technology | Trend extrapolation methods | Data source |
|---|---|---|---|---|
| You et al. (2017) | China, Canada | Coherent light generator | Autoregressive integrated moving averages (ARIMA), Bass model (S-shaped curve) | The United States Patent and Trademark Office (USPTO) website database |
| Zhang et al. (2017) | US | Concrete skyscraper technology | Logistic (Pearl) curve (S-shaped curve), Lotka–Volterra equations, Moore's law (exponential growth law) | The Global Tall Building Database of the CTBUH (2017) |
| Zhang et al. (2018) | US | Passenger airplane (system technology), turbofan aero-engine (component technology) | Logistic (Pearl) curve (S-shaped curve), Lotka–Volterra equations, Moore's law (exponential growth law) | Datasets from trade journals, Wikipedia, etc. |
| Xin et al. (2019) | China | 3,651 different technologies | Gated recurrent unit (GRU), Long short-term memory (LSTM), LSTNet, support vector regression (SVR), Conv-RNN (DNN as decoder), Conv-RNN-sep (multiple separate DNNs as decoder) | The United States Patent and Trademark Office (USPTO) website database, The UCR Time Series Classification Archive |
| Zhang et al. (2019a) | US | Central processing unit (CPU), passenger airplane | Lotka–Volterra equations, Moore's law (exponential growth law) | Wikipedia (CPU); 100 Years of Commercial Aviation, Aircraft Engines of the World 1970, and Jane's Aero-Engines (passenger airplane) |
| Zhang et al. (2019b) | US | Passenger airplane fuel efficiency | Logistic (Pearl) curve (S-shaped curve), Lotka–Volterra equations, Moore's law (exponential growth law) | Jane's All the World's Aircraft, Wikipedia |

| Authors | Country | Technology | Trend extrapolation methods | Data source |
|---|---|---|---|---|
| Sevilla and Riedel (2020) | UK, US | Quantum computing | Linear regression | Data reported in the previous 39 papers |
| Gui and Xu (2021) | China | Robotics | Autoregressive integrated moving averages (ARIMA), long short-term memory (LSTM), 2-layer-LSTM, EEMD-LSTM (EEMD: ensemble empirical mode decomposition) | SCI paper data from the Web of Science database |
| Meng et al. (2021) | UK, Italy | Energy technologies including nuclear electricity, onshore wind, offshore wind, PV module, alkaline electrolysis cells, proton exchange membrane electrolysis cells | Moore's law (exponential growth law), Wright's law (experience curve) | Datasets from multiple databases (e.g., International Renewable Energy Agency [IRENA]), research articles, and research organization reports. |
| Oliveira et al. (2021) | Brazil, Germany, Chile | Welding technologies including cold metal transfer (CMT) and gas metal arc welding (GMAW) | Autoregressive integrated moving averages (ARIMA), automatic ARIMA, simple exponential smoothing, Holt's exponential smoothing, Holt-Winters exponential smoothing, Baranyi model (S-shaped curve), Gompertz curve (S-shaped curve), logistic (Pearl) curve (S-shaped curve), exponential function (exponential growth law), Richards' curve (generalized logistic) (S-shaped curve) | Derwent Innovation Index (DII) database |

| Authors | Country | Technology | Trend extrapolation methods | Data source |
|---|---|---|---|---|
| Howell et al. (2021) | US | Space exploration technology | Autoregressive integrated moving averages (ARIMA), Autoregressive Integrated Moving Average with Exogenous variables (ARIMAX) | Datasets from multiple public sources |
| Ray et al. (2022) | India | Bacillus thuringiensis (Bt) technology | GM(1,1) grey model and its improved version | The Cotton Advisory Board of India |

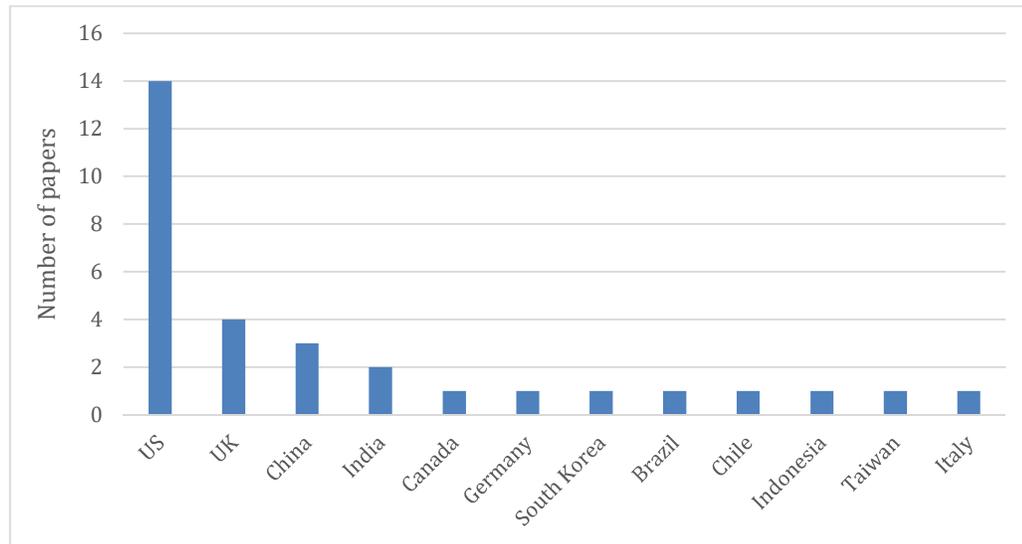

**Figure 2:** Number of papers and their origin countries

## 2.5 Interpreting the findings

In this step, the two research questions defined in the Introduction are addressed.

### 2.5.1 Research question one

*Q1: What does existing research provide about the implementation of quantitative trend extrapolation methods for technology forecasting?*

We discuss the 25 articles that were selected according to the indicated criteria next.

Inman et al. (2005, 2006) used two different techniques, multiple linear regression and TFDEA (technology forecasting using data envelopment analysis), to predict dates of the first flights of U.S. fighter aircraft. They took the 26 fighter jet data set for the period of 1944–1982 from a previous publication by Martino (1993) and used the 19 aircraft introduced between 1944 and 1960 to forecast the first flight dates of the jets between 1961 and 1982. Conclusions were then drawn based on their forecasting performance.

Yoon and Lee (2008) fit the Bass growth curve model to the patent application data collected from the United States Patent and Trademark Office (USPTO) database to forecast the future number of patents in nine categories of industries and compare the forecasting accuracy of each industry. For each industry, 35 time-series data points from 1965 to 1999 were retrieved. The model was trained on the first 30 data points between 1965 and 1994 and tested on the remaining 5 from 1995 to 1999 to compute the forecasting error. The results showed that, overall, the manufacturing sectors, such as the biotechnology and information technology sectors, worked better than the service sectors when applying patent-based technology forecasting.

Lamb et al. (2010) assembled airplane model data from aircraft interest websites, company websites, and relevant reports to forecast the first commercialization years of long-range passenger aircraft using simple and multiple linear regression models as well as the logistic (Pearl) curve. The data covered the first customer flight dates of 24 commercial airplane models over the period of 1965–2007. The first 16 models were chosen for training, while the latter eight were for model evaluation. It was found that two airplane parameters, maximum flight range and composite material percentage of the airplane structure, were significant predictors. In addition, the three technology forecasting methods appeared to be useful for forecasting passenger airplane technology commercialization.

Widodo et al. (2011) analyzed publication data using time series analysis to forecast trends in the frequency of technological terms dealing with apnea, such as movement disorders and cystic fibrosis, in biomedical technology. They queried the data from 1965–2010 using the PubMed database, resulting in 45 datasets of journal articles, where the first 35 datasets were selected for training and the latter 10 for testing. The models used included statistical time series models, such as Holt's exponential smoothing and Winter's exponential smoothing, and machine learning techniques, such as linear regression, polynomial regression, and support vector regression (SVR). The study found, after comparing the accuracy of the models, that machine learning techniques with the parameters properly initialized could achieve better performance than statistical time series models.

Yuan et al. (2011) utilized the GM(1,1) grey model, Bass model, and Gompertz curve to predict the power efficiency of white organic light-emitting diodes (OLED) and compare their prediction accuracy. The yearly power efficiency data from 1999 to 2006, a total of 10 observations, were collected and divided into model fitting (1999-2002) and testing (2003-2006) subsets. The study arrived at the conclusion that the GM(1,1) grey model had better forecasting performance than the Gompertz curve and Bass model.

Bailey et al. (2012) used the price and production data of sixty-two different technologies, the majority of which are associated with the energy, chemical, and information technology industries, from the Performance Curve Database, an online performance curves database (https://pcdb.santafe.edu/), to examine the relative forecasting accuracy of six different models across those technologies. The six models include Goddard's law, Moore's law, Wright's law, and three newly proposed models: Goddard's law random walk, Moore's law random walk, and Moore-Goddard's law random walk. The study found that production was a more accurate predictor of prices over longer time horizons than time, suggesting a Wright's law advantage compared to Moore's law.

Lim et al. (2012) applied the linear regression and TFDEA models to a data set of wireless communication technologies to forecast their commercialization years between 2001 and 2011. The data set was collected from the former research and extended to 4G mobile technologies for an updated analysis. It contained 23 records of different technologies, of which 19 were used for fitting and the remaining 4 for out-of-sample validation. The results indicated that the TFDEA model was likely to give more accurate predictions than the linear regression based on the metric of mean absolute deviation (MAD).

Sood et al. (2012) developed a new model, step and wait (SAW), and compared its predictive performance with that of traditional models, including Moore's law, Kryder's law, the logistic (Pearl) curve, the Bass model, the Gompertz curve, the Nave approach, the diff reg method, the Gupta model, and the tobit II model, for 25 technologies across six markets: automotive battery, data transfer, desktop memory, desktop printers, display monitors, and external lighting. In their work, each technology's most important attribute to customers was used as its performance measure. The data, primarily gathered from journals, press releases, and industry reports over several decades, were divided into two subsets for training and hold-out sample testing, for which the last five years of data were used. The analysis showed that the proposed SAW method, in comparison to the other approaches, had a lower prediction error.

Smith and Agrawal (2015) fit two time series forecasting models, Holt-Winters exponential smoothing (HWES) and autoregressive integrated moving average (ARIMA), to the patent data collected by the UC Berkley Fung Institute. Three patent groups, Database and File Management or Data Structures (DFMDS), Generic Control Systems or Specific Applications (GCSSA), and Artificial Intelligence (AI), from 1996 to 2013, were selected to be analyzed. The data set was split so that the validation set consisted of the last 15 months of data. Given the results, it was observed that all the models were able to yield acceptable forecasting results, with no one model performing best on all data sets.

Farmer and Lafond (2016) proposed a new model, the geometric random walk with drift, building on Moore's law, and applied it to data for 53 different technologies extracted from the Santa Fe Institute's Performance Curve DataBase (pcdb.santafe.edu). The model can give forecasts with prediction intervals for the cost improvement of a technology. Hindcasting, a form of cross validation, was used to fit and validate the model with a moving window approach based on the six most recent years of data for all technologies. To demonstrate the method, they made forecasts for solar photovoltaic (PV) modules and showed that the unit cost of a PV module is expected to continue decreasing at around a 10% annual rate, although there is a 5% chance that the price in 2030 will be higher than the price in 2013.

Nagula (2016) forecast the advancement of fuel cell technology in hybrid and electric vehicles using patent data and the Gompertz growth curve. The patent data associated with fuel cell technology from 1976 to 2012 came from the USPTO database. Three growth curve models were used in the analysis, and after comparing their forecasting performances, the Gompertz curve did better than the logistic (Pearl) curve and the Weibull model in forecasting fuel cell technology.

Cho and Anderson (2017) employed a data set of main battle tanks (MBTs) used in a previous paper and the methods of multiple linear regression and TFDEA to predict the MBTs' release year. The data set included 38 MBT models launched from 1941 to 1994, with 27 of them from 1941 to 1978 used for the training set and 11 from 1980 to 1994 for the hold-out test set. The study found that the linear regression model, using the four predictor variables of gun, horsepower, battle weight, and number of crew, best forecast the release year of MBT models with a mean absolute error (MAE) of 3.3 years.

You et al. (2017) applied the Bass and ARIMA models to study patent evolution for the technology of coherent light generators and compared both models' forecasting performance. They collected patent data for the period of 1976-2014 from the USPTO website. The first 372 months (from April 1976 to March 2007) of patent data were used for model fitting and the remaining 93 months (from April 2007 to December 2014) for testing. The results revealed that ARIMA performed better than Bass in forecasting the patent counts. More specifically, the models ARIMA(0,1,1) and ARIMA(0,2,3) achieved the best forecasting performance for the monthly and cumulative patent counts, respectively.

Zhang et al. (2017) fit the Lotka–Volterra model to concrete skyscraper height data during 1950–1996 from the Global Tall Building Database of the CTBUH (https://www.ctbuh.org) to forecast the development of concrete skyscraper technology. The prediction accuracy of the model on the hold-out sample was compared with that of the logistic curve and Moore's law. The authors concluded that the evolution of concrete skyscraper technology can be affected by the evolution of its component concrete technology, as the Lotka–Volterra model significantly improved prediction accuracy compared to Moore's law and the logistic curve.

Zhang et al. (2018) simplified the Lotka–Volterra equations and used the simplified equations to predict the performance of the passenger airplane (a system technology) and the turbofan aeroengine (a component technology). Their model took into account the interactions between the system technology and the component technology for modeling technology evolution. They collected the past performance data of the two technologies during 1960–2010 from various sources, such as trade journals, Wikipedia, etc. They fitted the model to the data during 1960–1998 and used the subsequent data as the hold-out sample to validate it. To assess the predictive performance of the simplified Lotka–Volterra equations, the logistic curve and Moore's law were also fitted to the same data for comparison. The results showed that the Lotka–Volterra based model outperformed the other models in predicting the hold-out sample.

Xin et al. (2019) devised two models, which they called Conv-RNN and Conv-RNN-sep, to forecast trends in patents using the USPTO patent data across different technologies. The patent data series was split into the training data and test data with a sliding window approach. The models were comprised of three layers, a convolution neural network (CNN) layer, a recurrent neural network (RNN) layer, and a deep neural network (DNN) layer. They were trained and tested and their performances on the test set were compared with those of other algorithms including the gated recurrent unit (GRU), long short-term memory (LSTM), support vector regression (SVR), and long- and short-term time-series network (LSTNet) algorithms. The results showed that both models had better performances in forecasting accuracy and consistency than the other models. It was found that the CNN layer contributed greatly to the two algorithms.

Zhang et al. (2019a) gathered CPU transistor count data during 1970-2018 from Wikipedia and fit Moore's Law to the data from 1970 to 2014 to forecast the hold-out samples from 2014 to 2018. The forecasts involved the prediction intervals that were generated with the bootstrap, or resampling, method. The authors also collected performance data of passenger airplanes from their previous publications during 1960-2008 and fit the Lotka–Volterra equations to the data from 1960 to 2004 to produce point forecasts and prediction intervals for 2004-2008. In these two cases, the hold-out testing revealed that their method successfully forecast the development of both technologies, as all the actual data in the hold-out samples were covered by the prediction intervals generated.

Zhang et al. (2019b) forecast passenger airplane performance using the Lotka–Volterra model and performance data during 1970-2017 compiled from Jane's All the World's Aircraft as well as Wikipedia. Their method took into account the interaction of the system, which is passenger airplane fuel efficiency, and its three components, which are aerodynamics, aero-engine fuel efficiency, and weight reduction as these components were found to significantly affect the fuel efficiency of passenger aircraft. The model was trained on the data during 1970-2000 and tested on the 2016 data as a hold-out sample. Their analysis showed that the performance of the Lotka–Volterra model exceeded that of the logistic curve models and Moore's law in prediction accuracy.

Sevilla and Riedel (2020) collected a dataset of quantum computer systems from 39 papers in the last two decades to develop an index, the generalized logical qubit (GLQ), that combines the two metrics widely reported in the literature, physical qubits and gate error rate, to measure and forecast progress in quantum computing technology. The trends in the best performance on the two metrics were extrapolated to predict the trend in GLQs using linear regression (LR) on the data in log space. The LR model used was validated using leave-one-out validation and rolling validation with different time spans. It was predicted that proof-of-concept fault-tolerant computation based on superconductor technology was unlikely (less than 5% chance) to exist before 2026 and that quantum devices capable of factoring RSA-2048 were unlikely (less than 5% chance) to be developed before 2039.

Gui and Xu (2021) constructed a combination model named EEMD-LSTM and applied it to publication data to forecast the development of robotics technology. The model's first layer, EEMD or Ensemble Empirical Mode

Decomposition, was used to break down the data into multiple components, and the second layer, LSTM, separately forecast the decomposed components. The forecasting results were then combined to form the final forecasts. The authors collected the robotics-related articles during 1996–2019 from the database of Web of Science and distributed the collected papers across eight primary areas of robotics study that were obtained through a text classification model. The yearly data on the number of papers in each research area were further split into the training data (from 1996 to 2016) as well as the test data (from 2017 to 2019). After comparing the forecasting performance of the proposed EEMD-LSTM on the test data with that of the ARIMA, LSTM, and 2-Layer LSTM models, they concluded that the model proposed was superior to the other methods in terms of lower mean and variance of the prediction errors, and thus deep learning models can perform better in technology forecasting.

Meng et al. (2021) utilized two types of approaches, expert elicitation methods and data-driven models (devised based on Moore's and Wright's laws), to generate technology cost forecasts for six different energy technologies and compared the forecasts against realized data to evaluate the methods' predictive power. The cost data and the data on installed capacity or cumulative production of the technologies over time used to carry out the estimation were obtained from various sources, such as the International Renewable Energy Agency, research articles, and research reports. The authors found that overall, the forecasted values from data-driven models were closer to the observed costs than those from expert elicitation methods.

However, all the methods used tended to predict slower cost reductions, implying a slower rate of technological improvement, than actually occurred in the technologies studied.

Oliveira et al. (2021) used the patent data from Derwent Innovation, a patent research platform, to derive the technology trend for two welding technologies, gas metal arc welding (GMAW) and cold metal transfer (CMT). Nine methods across the time series models and growth curves, namely ARIMA and Auto-ARIMA, Holt-Winters exponential smoothing, Holt's exponential smoothing, simple exponential smoothing, the Baranyi model, the Gompertz curve, the logistic (Pearl) curve, the exponential curve, and Richards' curve, were used and compared based on their forecasting performance for the last three years (2017-2019) of data for each technology. The findings indicated that among all the methods, Richards' curve gave the most accurate forecasts for CMT technology, while for GMAW technology, the forecasts from the Auto-ARIMA method were closest to the actual values. The study concluded that mathematical modeling could precisely predict the inflection points and the maturity phases of each technology.

Howell et al. (2021) applied ARIMA models and its extension, ARIMAX, to investigate improvements in space exploration technology and compared their forecasting performances. They collected data on spacecraft lifespan, which was used as a metric for improvement, for deep space missions from the beginning up to the present day. The data set was divided into two groups, with 85% used as a training set and 15% used as a test set. Their analysis concluded that ARIMAX models achieved better performance than ARIMA models in predicting improvements in spacecraft lifespan.

Ray et al. (2022) forecasted the development of Bacillus thuringiensis (Bt) technology in India using the yearly data for Bt cotton yield during 2002-2017 obtained from the Cotton Advisory Board of India. A GM(1,1) grey model and its improved version were fitted to the first eleven observations for training and tested on the remaining four for model evaluation. After comparing the forecasting accuracy of both models, they found that the improved GM(1,1) grey model performed better for forecasting the Bt cotton yield in India.

### 2.5.2 Research question two

*Q2: What are the quantitative trend extrapolation methods that have been adopted and what are the technologies where they have been applied?*

We grouped the quantitative trend extrapolation methods found in the selected studies into four main classes: growth curve, machine learning, time series, and other, which includes the methods not belonging to any of the other three classes. Table 4 lists the methods, method classes, authors of the articles that used these method classes, and the technologies each method was applied to. As can be seen from Figure 3, growth curves were the most used method class, followed by machine learning, time series, and other. The frequency of occurrence of the methods by classes across years is presented in Figure 4. It indicates that growth curves remain popular as a technology forecasting technique and, in particular, machine learning approaches have recently gained popularity with researchers and increased their share of the total from 2019 to 2022. Table 5 shows the breakdown of the methods used in the reviewed articles. The table contains information about the specific methods assigned to each method class, the number of articles using the methods, and their percentage of the total set of articles. Note that the sum of article counts exceeds the total number of the included articles because articles using multiple methods were listed multiple times.

**Table 4:** Summary of the method classes and forecasted technologies used in the included studies

| Method class | Method | Technology | Authors |
|---|---|---|---|
| Growth curve | Moore's law (exponential growth law) | 25 different technologies selected from six distinct markets: automotive motive battery technologies, data transfer, desktop printers, display monitors, desktop memory, and external lighting technologies | Sood et al. (2012) |
| | | Central processing unit (CPU), passenger airplane | Zhang et al. (2019a) |
| | | Concrete skyscraper technology | Zhang et al. (2017) |
| | | Energy technologies including nuclear electricity, onshore wind, offshore wind, PV module, alkaline electrolysis cells, proton exchange membrane electrolysis cells | Meng et al. (2021) |
| | | Passenger airplane | Zhang et al. (2018) |
| | | Passenger airplane fuel efficiency | Zhang et al. (2019b) |
| | | Sixty-two different technologies | Bailey et al. (2012) |
| | Logistic (Pearl) curve (S-shaped curve) | 25 different technologies selected from six distinct markets: automotive motive battery technologies, data transfer, desktop printers, display monitors, desktop memory, and external lighting technologies | Sood et al. (2012) |
| | | Commercial airplane | Lamb et al. (2010) |
| | | Concrete skyscraper technology | Zhang et al. (2017) |
| | | Passenger airplane | Zhang et al. (2018) |
| | | Passenger airplane fuel efficiency | Zhang et al. (2019b) |

| Method class | Method | Technology | Authors |
|---|---|---|---|
| | | Welding technologies including cold metal transfer (CMT) and gas metal arc welding (GMAW) | Oliveira et al. (2021) |
| | Bass model (S-shaped curve) | 25 different technologies selected from six distinct markets: automotive motive battery technologies, data transfer, desktop printers, display monitors, desktop memory, and external lighting technologies | Sood et al. (2012) |
| | | Coherent light generator | You et al. (2017) |
| | | Organic light-emitting diodes (OLEDs) | Yuan et al. (2011) |
| | | Various technologies by industries ( i.e., bio-technology, information technology, etc.) | Yoon and Lee (2008) |
| | Gompertz curve (S-shaped curve) | 25 different technologies selected from six distinct markets: automotive motive battery technologies, data transfer, desktop printers, display monitors, desktop memory, and external lighting technologies | Sood et al. (2012) |
| | | Fuel cell technology in hybrid and electric vehicles | Nagula (2016) |
| | | Organic light-emitting diodes (OLEDs) | Yuan et al. (2011) |
| | | Welding technologies including cold metal transfer (CMT) and gas metal arc welding (GMAW) | Oliveira et al. (2021) |
| | Wright's law (experience curve) | Energy technologies including nuclear electricity, onshore wind, offshore wind, PV module, alkaline electrolysis cells, proton exchange membrane electrolysis cells | Meng et al. (2021) |
| | | Sixty-two different technologies | Bailey et al. (2012) |

| Method class | Method | Technology | Authors |
|---|---|---|---|
| | Baranyi model (S-shaped curve) | Welding technologies including cold metal transfer (CMT) and gas metal arc welding (GMAW) | Oliveira et al. (2021) |
| | Exponential function (exponential growth law) | Welding technologies including cold metal transfer (CMT) and gas metal arc welding (GMAW) | Oliveira et al. (2021) |
| | Goddard's law (experience curve) | Sixty-two different technologies | Bailey et al. (2012) |
| | Kryder's law (exponential growth law) | 25 different technologies selected from six distinct markets: automotive motive battery technologies, data transfer, desktop printers, display monitors, desktop memory, and external lighting technologies | Sood et al. (2012) |
| | Richards' curve (generalized logistic) (S-shaped curve) | Welding technologies including cold metal transfer (CMT) and gas metal arc welding (GMAW) | Oliveira et al. (2021) |
| Machine Learning | Linear regression | Biomedical technology | Widodo et al. (2011) |
| | | Commercial airplane | Lamb et al. (2010) |
| | | Main battle tank (MBT) | Cho and Anderson (2017) |
| | | Quantum computing | Sevilla and Riedel (2020) |
| | | U.S. fighter jets | Inman et al. (2005, 2006) |
| | | Wireless communication | Lim et al. (2012) |
| | Long short-term memory (LSTM) | 3,651 different technologies | Xin et al. (2019) |
| | | Robotics | Gui and Xu (2021) |
| | Support vector regression (SVR) | 3,651 different technologies | Xin et al. (2019) |
| | | Biomedical technology | Widodo et al. (2011) |
| | 2-layer-LSTM | Robotics | Gui and Xu (2021) |

| Method class | Method | Technology | Authors |
|---|---|---|---|
| | Conv-RNN (DNN as decoder) | 3,651 different technologies | Xin et al. (2019) |
| | Conv-RNN-sep (multiple separate DNNs as Decoder) | 3,651 different technologies | Xin et al. (2019) |
| | Diff reg method (linear regression) | 25 different technologies selected from six distinct markets: automotive motive battery technologies, data transfer, desktop printers, display monitors, desktop memory, and external lighting technologies | Sood et al. (2012) |
| | Gated recurrent unit(GRU) | 3,651 different technologies | Xin et al. (2019) |
| | Long- and short-term Time-series network (LSTNet) | 3,651 different technologies | Xin et al. (2019) |
| | EEMD-LSTM (EEMD: ensemble empirical mode decomposition) | Robotics | Gui and Xu (2021) |
| | polynomial regression | Biomedical technology | Widodo et al. (2011) |
| Time Series | Autoregressive integrated moving averages (ARIMA) | Coherent light generator | You et al. (2017) |
| | | Data processing technology, including three classes, which are artificial intelligence (AI), database and file management or data structures (DFMDS), and generic control systems or specific applications (GCSSA). | Smith and Agrawal (2015) |
| | | Robotics | Gui and Xu (2021) |
| | | Space exploration technology | Howell et al. (2021) |
| | | Welding technologies including cold metal transfer (CMT) and gas metal arc welding (GMAW) | Oliveira et al. (2021) |
| | Holt-Winters exponential smoothing (Winter's exponential smoothing) | Biomedical technology | Widodo et al. (2011) |

| Method class | Method | Technology | Authors |
|---|---|---|---|
| | | Data processing technology, including three classes, which are artificial intelligence (AI), database and file management or data structures (DFMDS), and generic control systems or specific applications (GCSSA). | Smith and Agrawal (2015) |
| | | Welding technologies including cold metal transfer (CMT) and gas metal arc welding (GMAW) | Oliveira et al. (2021) |
| | Automatic ARIMA | Welding technologies including cold metal transfer (CMT) and gas metal arc welding (GMAW) | Oliveira et al. (2021) |
| | Geometric random walk with drift | 53 different technologies | Farmer and Lafond (2016) |
| | GM(1,1) grey model | Bacillus thuringiensis (Bt) technology | Ray et al. (2022) |
| | | Organic light-emitting diodes (OLEDs) | Yuan et al. (2011) |
| | Goddard's law random walk | Sixty-two different technologies | Bailey et al. (2012) |
| | Holt's exponential smoothing | Biomedical technology | Widodo et al. (2011) |
| | | Welding technologies including cold metal transfer (CMT) and gas metal arc welding (GMAW) | Oliveira et al. (2021) |
| | Moore's law random walk | Sixty-two different technologies | Bailey et al. (2012) |
| | Moore-Goddard's law random walk | Sixty-two different technologies | Bailey et al. (2012) |
| | Simple exponential smoothing | Welding technologies including cold metal transfer (CMT) and gas metal arc welding (GMAW) | Oliveira et al. (2021) |
| Other | Lotka–Volterra equations | Central processing unit (CPU), passenger airplane | Zhang et al. (2019a) |

| Method class | Method | Technology | Authors |
|---|---|---|---|
| | | Concrete skyscraper technology | Zhang et al. (2017) |
| | | Passenger airplane | Zhang et al. (2018) |
| | | Passenger airplane fuel efficiency | Zhang et al. (2019b) |
| | Gupta model | 25 different technologies selected from six distinct markets: automotive motive battery technologies, data transfer, desktop printers, display monitors, desktop memory, and external lighting technologies | Sood et al. (2012) |
| | Step and wait (SAW) model | 25 different technologies selected from six distinct markets: automotive motive battery technologies, data transfer, desktop printers, display monitors, desktop memory, and external lighting technologies | Sood et al. (2012) |
| | tobit II model | 25 different technologies selected from six distinct markets: automotive motive battery technologies, data transfer, desktop printers, display monitors, desktop memory, and external lighting technologies | Sood et al. (2012) |

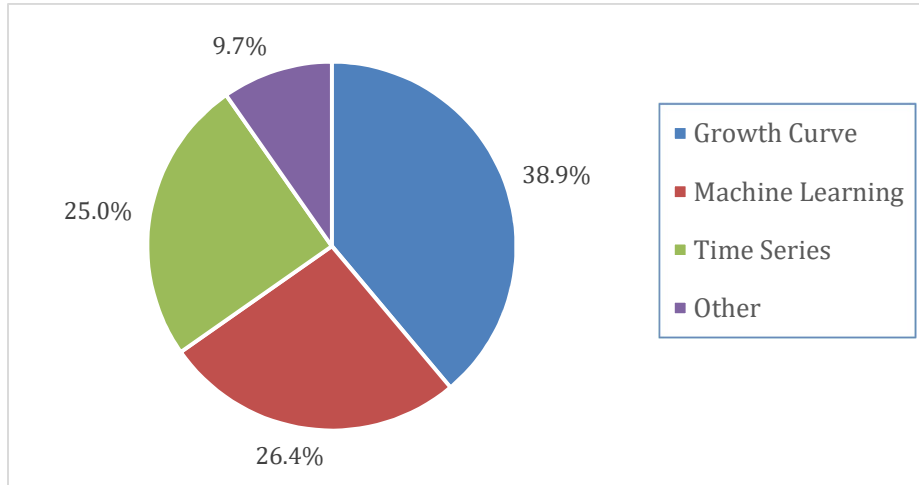

**Figure 3:** Distribution of method classes

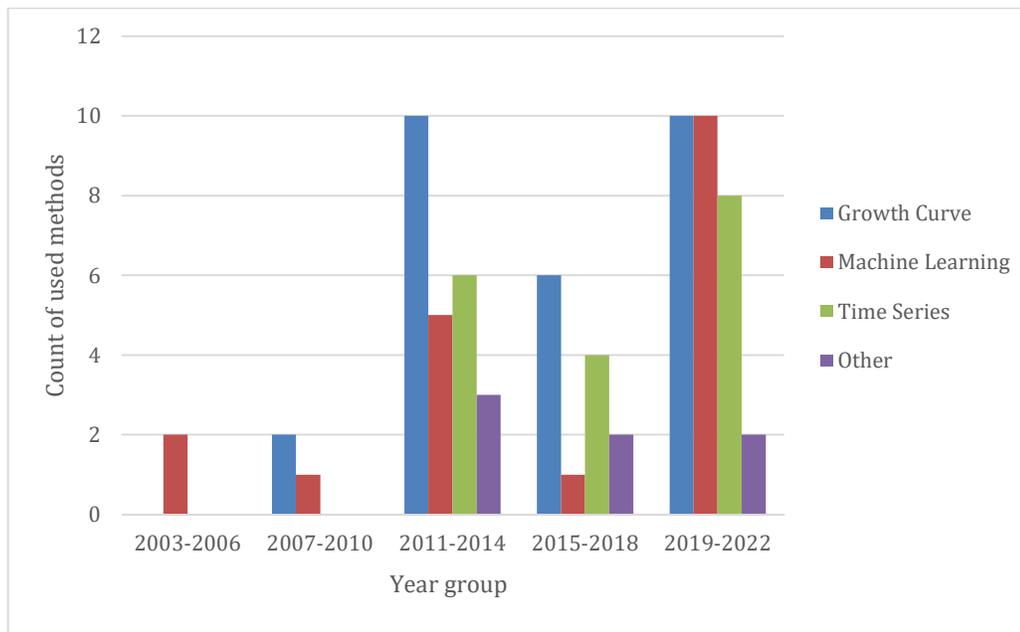

**Figure 4:** Distribution of method classes across the years

**Table 5:** Breakdown in the methods in terms of the number of their occurrences in the included articles

| Methods | | Article count | Percentage |
|---|---|---|---|
| **Growth Curve** | | **28** | **38.9%** |
| | S-shaped curve | 16 | 22.2% |
| | Exponential growth law | 9 | 12.5% |
| | Experience curve | 3 | 4.2% |
| **Machine Learning** | | **19** | **26.4%** |
| | Linear Regression | 9 | 12.5% |
| | LSTM | 3 | 4.2% |
| | SVR | 2 | 2.8% |
| | Conv-RNN | 1 | 1.4% |
| | Conv-RNN-sep | 1 | 1.4% |
| | EEMD-LSTM | 1 | 1.4% |
| | GRU | 1 | 1.4% |
| | LSTNet | 1 | 1.4% |
| **Time Series** | | **18** | **25.0%** |
| | Exponential smoothing | 6 | 8.3% |
| | ARIMA | 6 | 8.3% |
| | GM(1,1) grey model | 2 | 2.8% |
| | Geometric random walk with drift | 1 | 1.4% |
| | Goddard's law random walk | 1 | 1.4% |
| | Moore's law random walk | 1 | 1.4% |
| | Moore-Goddard's law random walk | 1 | 1.4% |
| **Other** | | **7** | **9.7%** |
| | Lotka–Volterra equations | 4 | 5.6% |
| | Gupta model | 1 | 1.4% |
| | Step and wait (SAW) model | 1 | 1.4% |
| | Tobit II model | 1 | 1.4% |

In the following, we review various quantitative trend extrapolation approaches presented in the studies.

- **Growth Curve**

Growth curves are frequently used in technology forecasting to model the behavior of product or technology performance over time (Cho and Daim, 2013). The growth curves used in the included articles can be classified into S-shaped curve (or S-curve), exponential growth law, and experience curve, based on the shapes of the curves. As shown in Table 5, the most widely used shape was the S-curves.

S-curves have long been used in biology to estimate measures such as population changes and bacterial growth over time. Researchers have recently extended their use to analyze technology life cycles and project the path of technological evolution. Similar to a product life cycle, an S-shaped curve pattern represents a pattern of growth that begins with slow development, followed by a rapid increase until reaching a maximum steepness at its inflection point, and subsequently a slowdown and return to minimal growth. S-curves encompass the logistic (Pearl) curve, Gompertz curve, Bass model, Richards' curve (generalized logistic), and many others, and can be classified into three different types based on the curve's behavior around the inflection point: symmetric, asymmetric, and flexible (Meade and Islam, 1998). Symmetrical S-curves, such as the logistic (Pearl) curve, have a fixed inflection point about which the rise and fall are symmetrical. Asymmetrical curves, such as Gompertz, also have a fixed inflection point but typically it is less than the midpoint between the upper and lower limits of growth. In flexible curves, such as the Weibull model, the inflection points are floating and some of them are less than the midpoint while others are greater.

Among different S-shaped curves, the most used two in technology forecasting are the logistic (Pearl) and the Gompertz (Martino, 2003). The equation for the logistic (Pearl) is given by:

$$Y_t = \frac{L}{1 + ae^{-bt}}.$$

The Gompertz is given by:

$$Y_t = Le^{-ae^{-bt}}.$$

In both, $Y_t$ is the performance measurement at time $t$, $L$ is the upper limit to the technological progress, $a$ and $b$, respectively, are the scale and shape parameters determining the location and shape of the curve, and $e$ = 2.718... is Euler's number. For further details about various S-shaped curves, refer to Meade and Islam (1998) and Young (1993).

The general form of the exponential growth law can be expressed as follows (Tague et al., 1981).

$$f(t) = ae^{bt}$$

where $f(t)$ represents the performance measurement at time $t$, $a$ and $b$ are constants, and $e$ is Euler's number.

One model that has a mathematically equivalent form as above and has been used extensively in the computer industry to predict the performance of computer chips is Moore's law, which emerged in 1965 when Gordon Moore, the co-founder of Intel and Fairchild Semiconductor, discovered that in the years between 1959 and 1965, the component count on cutting-edge integrated circuits had about doubled annually (Moore, 1965). He estimated that similar exponential progress would last for the next decade or longer. Over the years, a technology has been said to follow the generalized Moore's law when it improves exponentially with time (Sood et al., 2012; Nagy et al., 2013), and many researchers have applied the generalized Moore's law to study technological improvements for various technologies (Berleant et al., 2019; Farmer and Lafond, 2016; Magee et al., 2016; Nagy et al., 2013). A derivative of Moore's law, Kryder's law, is also an exponential growth law (Sood et al., 2012). It was proposed by Mark Kryder to predict a 13-month doubling period for storage capacity of computer disks (Walter, 2005).

The experience curve, also referred to as the learning curve, power law, and Wright's law, is an extensively used method in the literature to analyze the relationship between the cost of a technology and factors such as the cumulatively installed capacity or cumulative volume of production. It was originally observed by Wright (1936). Its basic form is as below (Alberth, 2008).

$$f(X) = aX^{-b}$$

In this equation, $f(X)$ represents the cost of a technology, $X$ can be the cumulative number of units produced or the capacity cumulatively installed, and $a$ and $b$ are constants. Another concept that has a similar power law relationship is Goddard's law, which was proposed by Goddard (1982) and uses annual production rather than cumulative production to explain the cost reduction of a technology.

The experience curve and exponential growth law have been broadly used in the literature. For instance, Bailey et al. (2012) fit Moore's law, Wright's law, and Goddard's law to the price and production data of sixty-two technologies mostly related to the energy, chemical, and information technology industries to forecast their performances. Also, Meng et al. (2021) utilized two types of approaches, expert elicitation methods and data-driven models (derived from Moore's and Wright's laws), to generate technology cost forecasts for six different energy technologies including nuclear electricity, onshore wind, offshore wind, PV module, alkaline electrolysis cells, and proton exchange membrane electrolysis cells.

- **Machine Learning**

Machine Learning (ML) is a data-driven approach for extracting complex relationships from a dataset. Various ML methods have been studied as alternatives to statistical modeling methods for forecasting (Makridakis et al., 2018). We expect more research applying this approach to technology forecasting problems as machine learning becomes more prevalent.

Linear regression is the most common form of regression analysis and a straightforward approach for supervised machine learning for modeling the relationship between a dependent (or response) variable and one or more independent (or predictor) variables. The earliest mathematical work on linear regression was attributed to Legendre and Gauss in the early nineteenth century, and since then, regression methods have remained an active research area and various transformations have been developed (Pierson and Gashler, 2017). Depending on the number of independent variables involved, linear regression can be classified into simple linear regression and multiple linear regression. The basic models for simple and multiple linear regression are expressed as follows (James et al., 2013).

Simple linear regression:

$$Y_i = \beta_0 + \beta_1 X_i + \epsilon_i$$

Multiple linear regression:

$$Y_i = \beta_0 + \beta_1 X_i$$

In both cases, $Y_i$ is the dependent variable, $X_i$ is the independent variable, $\beta_0$ is the intercept term, $\beta_1$, ..., $\beta_p$ represent parameters to be estimated for $p$ independent variables, the error term $\epsilon_i$ captures all other determinants of $Y_i$ other than the $X_i$, and the subscript $i$ indexes a particular observation. In addition to being used to identify relationships between variables, linear regression has been frequently carried out for forecasting. For example, Inman et al. (2005, 2006) used multiple linear regression to forecast the release year of U.S. fighter aircraft. Lamb et al. (2010) applied both simple and multiple linear regression to forecast the first commercialization years of long-range passenger aircraft. Cho and Anderson (2017) also used multiple linear regression to forecast the launch time of main battle tanks. Moreover, Dereli and Durmuşoğlu (2009, 2010a, 2010b) performed linear regression modeling of patent information to develop a trend-based patent alert system (PAS).

Polynomial regression is another form of regression analysis. In polynomial regression, the relationship between the dependent variable and the independent variable is modeled as an $n$th degree polynomial equation (James et al., 2013), which is generally expressed as

$$Y_i = \beta_0 + \beta_1 X_i$$

Polynomial regression is a type of regression producing a non-linear curve.

LSTM (Long Short-Term Memory), introduced by Hochreiter and Schmidhuber (1997), is a special type of recurrent neural network (RNN) that is able to not only model sequential events with standard RNNs but learn much longer-term temporal dependencies between events than standard RNNs, which suffer the loss of ability to link information as the time steps between events increase. It has been used in a large variety of tasks, such as machine translation (Sutskever et al., 2014), speech recognition (Graves et al., 2013), and technology forecasting (Gui and Xu, 2021). In an LSTM unit, a memory cell is introduced, in addition to the existing hidden state of RNNs, to keep information of past events in memory for long periods of time. Gating mechanisms are also employed to control the memory cell state and regulate how much of the previous information to forget and new information to take in before passing on the information to the next unit.

SVMs, or support vector machines, were developed by Vladimir Vapnik and his colleagues at AT&T Bell Laboratories in the 1990s. They are a machine learning technique that can handle tasks such as classification and regression analysis (Osuna et al., 1997). The idea of an SVM model is that an optimal separating hyperplane that has the largest gap between two classes in a higher-dimensional feature space is created based on training samples, and subsequent samples can be assigned to a class in that space depending on which side of the gap they map to. It was initially devised to resolve pattern classification problems and later extended to another version, which is called support vector regression (SVR), for dealing with the regression problems. The last two decades have seen a growing interest in the SVM approach among researchers and various applications can be found in the literature. More details about SVMs and their application can be found in Osuna et al. (1997).

The Conv-RNN and Conv-RNN-sep models were proposed by Xin et al. (2019) to forecast trends in patents. The architecture of the models comprises a CNN layer as a feature extractor to extract patterns from the data and an RNN as an encoder to encode a sequence of input features into an intermediate representation. Also, the Conv-RNN model has a DNN as a decoder that decodes the representation into the forecasting results whereas the Conv-RNN-sep uses multiple separate DNNs. Xin et al. (2019) argued that adding the CNN layer helps improve the forecasting performance of the models as it captures the patterns in the data.

The EEMD-LSTM model is composed of two models: EEMD and LSTM. EEMD (ensemble EMD) is an improvement to empirical mode decomposition (EMD), which was proposed by Huang et al. (1998) to study non-linear and non-stationary behavior in a complex signal. EMD/EEMD performs decomposition by decomposing a signal into multiple components, known as intrinsic mode functions (IMFs), with different amplitudes and frequencies and a residue (Gui and Xu, 2021). Wu and Huang (2009) developed the EEMD algorithm to resolve the mode mixing phenomenon problem while preserving the non-linear and non-stationary features of the original sequence data. The problem is defined as a single IMF containing signals of different scales, or different IMFs containing a signal of a similar scale, and is caused by intermittence or noise signals in the EMD algorithm.

GRU (gated recurrent unit), proposed by Cho et al. (2014), is a variant of LSTM designed to have fewer gates and parameters than LSTM due to the high computational cost of LSTM. A GRU has a reset gate and an update gate but not the output gates that are present in LSTM to control the output flow of memory information into the future hidden state. Additionally, a GRU does not have a separate memory cell but uses the gating mechanism to keep and control the memory information within the unit. Despite using simplified gating mechanisms, GRU still has comparable result quality to LSTM according to empirical evaluation (Chung et al., 2014). Therefore, GRU can be used when less memory is available and faster computation desired.

LSTNet, or long- and short time-series network, is a deep learning model introduced by Lai et al. (2018) for multivariate time series forecasting. It was designed to model the mixture of both long-term and short-term repeating patterns in time series data. In LSTNet, a convolutional neural network (CNN) is utilized to capture short-term dependency patterns among variables, which are fed simultaneously into a recurrent neural network, a temporal attention layer, and an autoregressive component to discover long-term patterns in data.

- **Time Series**

A time series is typically a succession of data points observed at equally spaced periods of time. Time series forecasting refers to the process of forecasting future values of a variable with a mathematical model based on previously observations. Time series forecasting models have been used extensively in both scientific and business fields (Smith and Agrawal, 2015), although they are not frequently employed in forecasting technological advancement (Zhang et al., 2019a). They range from simple exponential smoothing methods to more complex ones, such as the method of autoregressive integrated moving average (ARIMA).

The exponential smoothing methods originated in the 1950's and 1960's from the work of Brown (1959), Holt (1957, reprinted 2004), and Winters (1960) when they were developing forecasting models for inventory control (Fomby, 2008). The methods have been used in various fields especially in forecasting economic and financial data. Exponential smoothing works by smoothing a time series by allocating unequal weights to the previous observations and using the smoothed series to forecast near-future values of the series. Older observations are given less weight than recent observations and the weights decrease exponentially towards zero as the distance from the present increases. In other words, exponential smoothing relies more on recent observations than on older observations to forecast. The idea is that recent observations are often the best forecasters of the future.

The most basic exponential smoothing method is simple exponential smoothing, which requires little computation and is suitable for forecasting a data series that moves randomly around the mean (or level) of the series without a clear trend or seasonality. Simple exponential smoothing is given by the equation (Nau, 2014)

$$S_t = \alpha * X_t + \left( 1 - \alpha \right) * S_{t-1}.$$

Here, $X_t$ is the actual value or observation at current time step $t$, $S_t$ is the current smoothed value or level estimate and the forecast for the next time step $t+1$, $S_{t-1}$ is the previous level estimate, and $\alpha$ is the smoothing parameter to be determined. The equation is recursive since every level estimate is calculated using every estimate before it. The computation eventually leads to a weighted average of past observations across all time steps with the weights determined by the value of $\alpha$, which is set between 0 and 1. If $\alpha$ is set to near one, the smoothed value or next forecast $S_t$ will more closely resemble the current observation. If it is near zero, $S_t$ will appear very similar to the old forecast.

Simple exponential smoothing can produce good results for a time series with a nearly horizontal level, but it will result in bad forecasts in case of existence of a trend. To overcome this deficiency, Holt's exponential smoothing (Holt, 1957, reprinted 2004), also known as double exponential smoothing, incorporates a trend term to allow the forecasting of data with a trend. In the Holt's method, two smoothing equations are used to respectively estimate the level and trend of the series at the end of each period. In the level equation, the estimated level of the series at time $t$ is a weighted average of the actual value of the series at time $t$ and the one-step-ahead forecast for time $t$. In the trend equation, the estimated trend of the series at time $t$ is a weighted average of the difference of the estimated level of the series at time $t$ and the estimated level of the series at time $t$-1 and the estimated trend of the series at time $t$-1. Both the estimates of the level and trend of the series are then used to generate the $h$-step-ahead forecast, which equals the estimated level plus the estimated trend value multiplied by $h$, where $h$ represents the number of time steps into the future.

Holt-Winters exponential smoothing, also called Winters' exponential smoothing or triple exponential smoothing, improves Holt's exponential smoothing by capturing seasonality on top of the trend to cope with data series with both trend and seasonal patterns, which reflect a tendency to exhibit a pattern that repeats itself at fixed intervals. The Holt-Winters method, therefore, consists of three smoothing equations including the level, the trend, and the seasonality update equations. In general, depending on the form of seasonality, there are two variations of the Holt-Winters method: the multiplicative method which assumes that the seasonal fluctuations vary proportionately to the data series' mean level and the additive method which assumes that the seasonal fluctuations are stable (Chatfield, 1978). The method used must correspond to the particular type of time series. When applying either the additive or multiplicative method to a series, initial estimates for the level, trend, and seasonal components must be determined. As more observations become available, the values for the three components will all be updated with the corresponding smoothing parameters. Subsequently, forecasts for any number of steps forward can be made. Chatfield (1978) made some practical suggestions regarding the implementation of the Holt-Winters approach. Goodwin (2010) also provided an easy-to-follow example illustrating this method of exponential smoothing. For detailed discussion of both the Holt's and Holt-Winters methods, see Kalekar (2004).

The autoregressive integrated moving average (ARIMA) models (also known as Box–Jenkins models) were introduced by Box and Jenkins (1970, 1976) to integrate the autoregressive (AR) model, in which the past values of a time series can determine its current value, and the moving average (MA) models, in which past forecast errors help determine its current value, to allow for modeling non-stationary time series data (De Gooijer and Hyndman, 2006). A non-stationary time series, which has a mean that changes over time as opposed to remaining roughly constant, can be identified with a visual inspection of the plot of the series or through the augmented Dickey-Fuller test to check whether the series is non-stationary. If a time series is determined to be non-stationary, it can be made stationary by repeatedly taking the difference of the series until it becomes stationary. However, caution should be taken to avoid over-differencing, which would produce a loss in forecasting performance (Hyndman and Khandakar, 2008). Differencing is a discrete operation similar to differentiation, and it tends to magnify fluctuations due to noise. Multiple differencing stages tend to lead to noise that overwhelms the underlying signal of interest.

The ARIMA models are typically notated as ARIMA(p,d,q), in which p is the order of the autoregressive (AR) process, d represents the order of differencing required for making the original non-stationary time series stationary, and q is the order of the moving average (MA) process (Demos and Salas, 2017). Selecting appropriate values for p, d, and q is not an easy task but can be automatically done by software implementing automatic ARIMA (Auto-ARIMA) forecasting (Hyndman and Khandakar, 2008). The best possible ARIMA model can also be obtained with the aid of software using likelihood-based model fit measures, for example, the Akaike's Information Criterion (AIC) and the Bayesian Information Criterion (BIC) (De Gooijer and Hyndman, 2006; Howell et al., 2021). A detailed discussion of ARIMA models can be found in Demos and Salas (2017). Smith and Agrawal (2015) studied patent data and found the results satisfactory using ARIMA(1,1,3), ARIMA(2,1,0), and ARIMA(2,1,1) to forecast the patent counts of three patent groups respectively.

The geometric random walk with drift model was proposed by Farmer and Lafond (2016) based on Moore's law for making distributional forecasts for a technology that clearly account for forecast uncertainty rather than being limited to point forecasts. Applying it to solar photovoltaic modules, Farmer and Lafond (2016) showed that the model can be used to forecast technological cost improvement, although it is not suitable for unevenly spaced time series data (Zhang et al., 2019a).

The GM(1,1) model, expressing a first order differential equation with one input variable, is a grey forecasting model based on the grey system theory that was proposed in 1980s by Deng (Yuan et al., 2011). Since then, it has been broadly applied to various systems in domains such as economics, engineering, finance, geology, medicine, social systems, transportation, etc. (Kayacan et al., 2010) and recently expanded to address forecasting technological progress (Martino, 2010). The "grey" in the grey system theory implies that the theory deals with the systems characterized by both 'black' or unknown information, 'white' or known information, and 'grey' or partially known information, exemplified by grey numbers of which an interval is a typical example. In other words, grey forecasting models require only a limited amount of data to estimate the behavior of the unknown system and are thus suitable for forecasting the performance of a new technology with limited historical data. Empirical evidence has shown that grey modeling performs better in short-term forecasting than in medium- and long-term forecasting (Hsu, 2003). However, it is typically true that short-term forecasting in general works better than medium- and long-term forecasting. Yuan et al. (2011) used the GM(1,1) model to predict the power efficiency of OLEDs and found it more accurate than two S-shaped curves, namely the Gompertz curve and the Bass model. Lin et al. (2004) provided an explanation of the grey models.

- **Other**

This class includes four methods: Lotka–Volterra equations, the Gupta model, the step and wait (SAW) model, and the tobit II model.

The Lotka–Volterra equations were originally devised in the early 1900s to explain changes in fish and shark populations observed in the Adriatic Sea and have been frequently used to analyze predator-prey competition systems (Zhang et al., 2018). The method's use in modeling the dynamics of interacting populations in an ecosystem has also been extended to model other relationships (Zhang et al., 2019b), such as multi-technology interactions. In 1997, the original Lotka–Volterra equations were amended by Pistorius and Utterback (1997) to study the interaction between two technologies (Zhang et al., 2019b). Their method consists of a pair of differential equations where the left-hand side contains the derivatives of the performance measures of two technologies, which represent the two technologies' performance change rates over time, respectively, and the right-hand side contains the variables that represent the performance measures of the two technologies and the interaction terms that represent the interaction effects between the two technologies (Zhang et al., 2018). The equations can be easily augmented to include more technologies to model technology evolution and interaction (Dohnal and Doubravsky, 2016).

The Gupta model is widely used in the marketing literature to analyze sales promotion effects and predict consumers' interpurchase behavior. Its discontinuous nature also makes it suitable for modeling the irregular course of technological change. The model can be decomposed into three stages: brand choice, interpurchase time, and purchase quantity. Sood et al. (2012) fit the purchase quantity model and interpurchase time model to predict the performance improvement of a technology and the time between changes in performance, respectively.

The step and wait (SAW) model was introduced by Sood et al. (2012) to model the technologies displaying an irregular growth pattern that contains periods of constant performance (flat waiting periods between steps) and discontinuous steps. In this method, the change in technological performance is modeled by the step sub-model and the time between changes is modeled by the wait sub-model. The authors argued that this discontinuous model is better at capturing technological improvement processes than some traditional continuous approaches such as exponential curves and S-shaped curves.

The tobit II model, or the tobit type II, is one of the tobit model variations that Goldberger (1964) named in reference to Tobin (1958), who developed the first tobit model to address issues with limited dependent variables (Tellis, 1988), which are dependent variables that have limited ranges with many observations at the lower or upper limit (Econometrics Academy, 2022). With tobit II, technology evolution can be modeled with a set of step functions (Sood

et al., 2012).

## 3. Discussion and conclusions

As technology forecasting has continued to attract attention, technology forecasting techniques have been rapidly developed. We sought to review the existing literature on technology forecasting addressing the application of quantitative trend extrapolation approaches to various technologies. Growth curves and time series methods were shown to continue to play a significant role in technology forecasting, while some new methods, such as machine learning approaches, including machine learning-based hybrid models, have also emerged in recent years. Several studies in our review have shown hybrid models producing more accurate forecasts than individual models. In this regard, hybrid models seem to have great potential for improving forecasting accuracy and show a promising venue for future research. Despite the excitement surrounding these new developments, more evidence is needed before machine learning and hybrid models can be concluded to be better than growth curves and traditional time series methods. Therefore, we expect the growing trend of developing and applying hybrid models to technology forecasting to continue.

Our study has limitations that provide opportunities for future literature reviews. First, our search was restricted to the publications retrieved from the Google Scholar and Web of Science databases. There are, however, many other well-known databases, such as Scopus, SpringerLink, and IEEE Xplore, that were not employed in this survey but might contain additional relevant studies. In future reviews, these databases could be included to reduce the chances of overlooking a relevant study. Second, by limiting our review to the articles published during 1990-2022, older articles were not included. Future review activities could expand the time span to also cover such studies, illuminating the longer term history of the field. Nevertheless, the results of this study provide technology forecasting researchers and practitioners with current knowledge of the area and a perspective on commonly used quantitative technology forecasting techniques.